%% file: main.tex
\documentclass[runningheads]{llncs}

\usepackage[table]{xcolor} % to color rows
\usepackage{graphicx}
\usepackage{amsmath}
\usepackage{subfig}
\usepackage[colorlinks]{hyperref}
\usepackage[linesnumbered,ruled]{algorithm2e} % for code
\usepackage{tikz}
\usepackage{multirow} % for tables
\usepackage{amssymb}
\usepackage{pifont} % for cross sign
\usepackage{float}
\usepackage{booktabs}
\usepackage{microtype}

\usepackage[makeroom]{cancel} % for table
\def\checkmark{\tikz\fill[scale=0.4](0,.35) -- (.25,0) -- (1,.7) -- (.25,.15) -- cycle;}

\begin{document}
\title{Extreme Consistency: A Simple, Generic, and Versatile Semi-supervised Learning Framework for Handling Limited Data and Domain Shifts}

\title{Extreme Consistency: Overcoming Annotation Scarcity and Domain Shifts}

% \title{Overcoming Annotation Limitation and Domain Shifts: Extreme Consistency to the Rescue}
\author{Gaurav Fotedar$^*$, Nima Tajbakhsh$^*$, Shilpa Ananth, and Xiaowei Ding}
\authorrunning{***}
% \institute{***}
\institute{VoxelCloud Inc.\\
\email{\{gauravfo, ntajbakhsh, shilpa, xding\}@voxelcloud.io}\\
%\url{***}
}
\maketitle              % typeset the header of the contribution

\begin{abstract}
Supervised learning has proved effective for medical image analysis. However, it can utilize only the small labeled portion of data; it fails to leverage the large amounts of unlabeled data that is often available in medical image datasets. Supervised models are further handicapped by domain shifts, when the labeled dataset, despite being large enough, fails to cover different protocols or ethnicities. In this paper, we introduce \emph{extreme consistency}, which overcomes the above limitations, by maximally leveraging unlabeled data from the same or a different domain in a teacher-student semi-supervised paradigm. Extreme consistency is the process of sending an extreme transformation of a given image to the student network and then constraining its prediction to be consistent with the teacher network's prediction for the untransformed image. 
% Extreme consistency improves the student model through strong regularization based on the unlabeled data, which, otherwise, would  have remained unused during training. 
The extreme nature of our consistency loss distinguishes our method from related works that yield suboptimal performance by exercising only mild prediction consistency. Our method is 1) auto-didactic, as it requires no extra expert annotations; 2) versatile, as it handles both domain shift and limited annotation problems; 3) generic, as it is readily applicable to classification, segmentation, and detection tasks; and 4) simple to implement, as it requires no adversarial training. We evaluate our method for the tasks of lesion and retinal vessel segmentation in skin and fundus images. Our experiments demonstrate a significant performance gain over both modern supervised networks and recent semi-supervised models. This performance is attributed to the strong regularization enforced by extreme consistency, which enables the student network to learn how to handle extreme variants of both labeled and unlabeled images. This enhances the network's ability to tackle the inevitable same- and cross-domain data variability during inference.
\end{abstract}


\section{Introduction}

Deep learning has proved effective in automating the diagnosis and quantification of various diseases and conditions. However, the vast majority of such models have limited clinical value, because they are trained and evaluated on labeled datasets that lack adequate annotation or show poor data diversity. The former problem is mainly caused by limited annotation budgets, while the latter, which causes the domain shift problem, arises when a dataset has poor coverage of ethnicities, imaging protocols, and scanner devices.  In practice, securing a large, diverse labeled dataset is hardly feasible if not impossible; therefore, it is critical to devise learning strategies that increase model robustness against such dataset limitations through the effective utilization of unlabeled data.

To mitigate both limited annotation and domain shift problems, we propose a Semi-Supervised Learning (SSL) method, called \emph{Extreme Consistency}. The main idea behind extreme consistency is to regularize the model by forcing predictions for the original data and its extreme transformations to be consistent. This idea is fundamentally different from data augmentation (Table~\ref{tab:ablation}(c)), because the latter can be applied only to labeled data and, further, lacks explicit consistency constraints. We realize our method through a teacher-student paradigm, where we feed an extreme variant of a given image to the student network and constrain its prediction to be consistent with the teacher's prediction for the original image.  Importantly, extreme consistency relies on a diverse set of intensity-based, geometric, and image mixing transformations, which are blended based on a prior probability. If the combined transformation consists of only intensity manipulations, consistency loss reduces to an invariance constraint; that is, the predictions for the original and transformed image must be identical. However, inclusion of geometric and mixing augmentation will elevate this to an equivariance constraint; i.e., predictions must exhibit the same geometric transformations applied to the original images. Owing to the stochastic nature of diverse transformations, extreme consistency effectively spawns numerous invariance and equivariance constraints, providing strong regularization and thus higher-performing models. Critically, we exclusively apply diverse transformations to the inputs of the student network; thus, the teacher will receive ``easy" examples that have only mild augmentation, which facilitates a healthy reference prediction for the consistency loss. Our experimental results on three public datasets demonstrate that the exclusive and diverse transformations make critical contributions to the success of our method, enabling a new level of performance that surpasses recent semi-supervised models  \cite{li2019transformation,cui2019semi,perone2019unsupervised,dong2018unsupervised}, as well as supervised models based on U-Net and UNet++ \cite{zhou2018unet++}. 

Extreme consistency offers the following advantages: First, our method is auto-didactic, because extreme consistency uses the teacher's predictions on unlabeled data, requiring no additional expert annotations. Second, our method is versatile, handling both domain shift 
%(where unlabeled data comes from a domain other than the labeled set)
and limited annotation problems.
%(where the labeled set is small, and unlabeled and labeled data belong to the same domain)
This is because extreme consistency exposes the model to extreme variants of labeled and unlabeled data, which enhances the capability to cope with future same- and cross-domain data disparity. Third, our method is generic because extreme consistency can be applied to classification, segmentation, and detection tasks by subjecting logits, masks, and saliency maps to the extreme consistency loss, respectively. Note that saliency or center maps are used in recent object detectors such as CenterNet \cite{duan2019centernet}. Fourth, our method is simple to implement, requiring neither adversarial training nor any changes in the architecture of the base model.

{\bf \underline{Our contributions}} include 1) a novel semi-supervised approach based on extreme consistency, 2) a cross-dataset image mixing transformation to expand the unlabeled data for stronger consistency regularization, 3) a comprehensive ablation study on the individual components of our method, 4) extensive evaluation using three public datasets for the problems of limited annotations and domain shifts, and 5) demonstrated gains over several semi-supervised baselines as well as supervised baselines trained only with labeled data.

\section{Related Work}

% Semi-supervised learning is a training paradigm wherein both labeled and unlabeled data are used to train the model for a target task. This paradigm is also referred to as unsupervised domain adaptation, when the unlabeled data comes from a domain other than that of the labeled dataset. 
In this section, we briefly review the semi-supervised learning and unsupervised domain adaptation methods that are most relevant to our work, and refer the readers to \cite{tajbakhsh2019embracing} for a comprehensive review of these methodologies.

% Semi-supervised learning has proved effective in handling limited data. 
% The existing SSL methods differ in how they leverage unlabeled data during the course of training. 
Early SSL methods were based on adversarial zero-sum games, which, at their equilibrium, generate similar feature embedding \cite{sedai2017semi,baur2017semi} or similar predictions \cite{zhang2017deep,mondal2018few} for labeled and unlabeled data. 
%However, these methods are difficult to train due to their sophisticated adversarial architectures  and  numerous sensitive  hyper-parameters. In contrast, our method is straightforward to train, as it does not require any adversarial training. Our method is further easy to tune as it has only a few hyper-parameters to tune.
Recent SSL methods have, however, strived for simplicity and practicality. One of the popular recent themes is to train the model so it generates consistent results for a given image undergoing certain transformations. The consistency loss in \cite{cui2019semi,yu2019uncertainty} encourages invariance against only noise perturbations, while the consistency loss in \cite{bortsova2019semi,li2019transformation} encourages equivariance against geometric transformations. Finally, the consistency loss in \cite{perone2019unsupervised} is similar to \cite{bortsova2019semi,li2019transformation}, but benefits from a teacher-student paradigm. Our method is also based on consistency loss; however, it is not limited to certain geometric transformations (flip and rotation in \cite{li2019transformation}) or certain types of additive or multiplicative noise perturbations (Gaussian noise in \cite{cui2019semi,yu2019uncertainty}). This is because specific transformations may only be effective for a limited range of applications. For instance, consistency under perturbation may not be as effective in ultrasound images or optical coherence tomography scans that have inherent speckle noise or low signal-to-noise ratio. 
%We overcome this limitation by including a diverse set of intensity, geometric, and image mixing transformations, resulting in numerous invariance \emph{and} equivariance constraints imposed through the consistency loss. Our method thus generates stronger regularization and richer supervision signals, which, as we will show in Section~\ref{sec:exp}, leads to consistently superior performance across tasks and datasets. 
A concurrent work similar to ours is the arXiv submission~\cite{sohn2020fixmatch}, which considers only image classification and tackles the limited annotation problem in the context of natural images. By contrast, our paper tackles image segmentation in medical images, benefits from image mixing augmentation, and targets both domain shift and scare annotation problems.

The domain adaptation methods fall into two major categories. The first consists of methods that convert labeled source images such that they take the appearance and style of the target domain images \cite{huo2018synseg,chen2019synergistic}, whereas the second converts the target domain images into the source  domain images \cite{chen2018semantic,Giger2018}. Despite the differences, methodologies in both categories heavily rely on different variants of adversarial networks. By contrast, our approach is easy to train, requiring no sophisticated adversarial training.

\section{Method}

Our semi-supervised learning method follows the teacher-student paradigm. Specifically, our architecture consists of a student network $N_s$, whose weights are updated through backpropagation, and a teacher network $N_t$ whose weights simply track the weights of the student network through an exponential moving average filter. During training, the student network has two learning objectives: 1) minimizing the target supervised loss for the labeled data, 2) minimizing an unsupervised  consistency loss by generating predictions that are consistent with the reference predictions generated by the teacher network. The novel aspect of our method is the notion of extreme consistency, which maximally utilizes unlabeled data during training.
% we realize through a diverse yet exclusive set of image transformations.

\vspace{4pt}
\noindent{\bf Extreme consistency:} Extreme consistency is the process of feeding an extreme variant of a given image to the student network and constraining its prediction to be consistent with the teacher's prediction for the original image. The extreme nature of our consistency loss distinguishes our method from similar works (e.g., \cite{li2019transformation,yu2019uncertainty,cui2019semi}) that exercise prediction consistency under a limited set of transformations. We take the prediction consistency to extremes through a diverse set of transformations, which effectively spawns numerous invariance and equivariance constraints on the student network. Invariance constraints are imposed when the two networks receive variants of the same input that differ only in intensity values with no change in the spatial layout; otherwise, the applied transformation leads to an equivariance constraint; that is, the resulting predictions from the two networks must reflect the same geometric differences as do the original and transformed images. We hypothesize that extreme consistency provides stronger regularization, thereby enabling a higher-performing student network.

\vspace{4pt}
\noindent{\bf Diverse transformations:} We generate diverse variants of an input image through an extreme image transformation module $T_e$.
For this purpose, we propose an image mixing augmentation, which, when combined with other intensity manipulations (e.g., posterization, equalization, etc.) and geometric transformations (rotation, scale, and translation), can effectively generate a diverse set of out-of-distribution images. As discussed in \cite{dai2017good}, these extremely distorted images can potentially improve the performance of an SSL method. We briefly explain our image mixing methods and refer the reader to our supplementary material for the detailed composition of the $T_e$ module.

\vspace{4pt}
\noindent{\bf Cross-dataset image mixing:}
Our image mixing technique aims to expand the unlabeled dataset by cutting a random region from a labeled image and pasting it into an unlabeled image. The prediction for a mixed image is compared for consistency against the prediction for the labeled image only in the pasted region, and against the prediction result for the unlabeled image in all other pixels. Our image mixing method is inspired by \cite{yun2019cutmix,french2019consistency}, but differs from \cite{yun2019cutmix}, which is data augmentation for image classification, and from \cite{french2019consistency}, which mixes only same-domain images. 

\vspace{4pt}
\noindent{\bf Exclusive transformations:} Applying diverse transformations to the inputs of both student and teacher networks, however, either leads to model divergence during training or results in only a marginal improvement over a supervised baseline. We have experimentally determined that it is critical to use the diverse transformations exclusively for the inputs to the student network and apply only mild transformations to the inputs of the teacher network. Intuitively, this design makes sense because the success of the consistency loss, especially for unlabeled data, relies on a healthy reference prediction, which in our case is generated by the teacher network. It is therefore necessary to feed ``easy'' examples to the teacher while sending the ``hard'' examples to the student. Omitting this critical design choice may have shifted previous works towards consistency based on just a narrow set of transformations, leading to sub-optimal improvements. 

\begin{figure}[t!]
\begin{minipage}[t]{0.65\textwidth}
  \vspace{0pt}  
   \begin{algorithm}[H]
  \scriptsize
    \SetKwInOut{Input}{Input}
    \SetKwInOut{Output}{Output}

    % \underline{function Euclid} $(a,b)$\;
    \Input{labeled data $\mathcal{L}$, unlabeled data $\mathcal{U}$, student ($N_s$) and teacher ($N_t$) networks, consistency loss weight $\lambda$, moving average decay $\alpha$}
    \Output{Segmentation network}
    \For {$t\leftarrow 1$ \KwTo $\mathcal{T}$}{
        \tcc{sample labeled and unlabeled data}
        $\{X^l,Y\} \leftarrow sample(\mathcal{L})$; $ \{X^u\} \leftarrow sample(\mathcal{U})$\;
        
        \tcc{instantiate transformations}
        $T_b, T_e \leftarrow \mathcal{T}_b, \mathcal{T}_e$\;
        %; \quad \{X^u\} \leftarrow T_B(\{X^u\})$\
        
        \tcc{compute segmentation loss}
        $L_{seg} \leftarrow Dice(N_s(T_b(X^l)), T_b(Y))$\;
        
        \tcc{compute extreme consistency loss}
        $L^l_{ec} \leftarrow sDice(T_e(N_t(T_b(X^l))), N_s(T_e(T_b(X^l))))$\;
        $L^u_{ec} \leftarrow sDice(T_e(N_t(X^u)), N_s(T_e(X^u)))$\;
        
        \tcc{update student network}
        $L = L_{seg} + \lambda (L^l_{ec} + L^u_{ec})$;
        $minimize(L)$\;
        
        \tcc{update teacher network}
        \For {($w_t,w_s$) in ($N_t$,$N_s$)}{
        $w_t = \alpha\times w_t + (1-\alpha)\times w_s$\;
        }
        
    }
    
    \Return $N_t$
    \caption{}%Pseudocode for extreme consistency
    \label{algo:ssl}
\end{algorithm}
\end{minipage}%
~~~~\begin{minipage}[t]{0.35\textwidth}
\vspace{0pt}  
\includegraphics[width=0.99\linewidth, keepaspectratio]{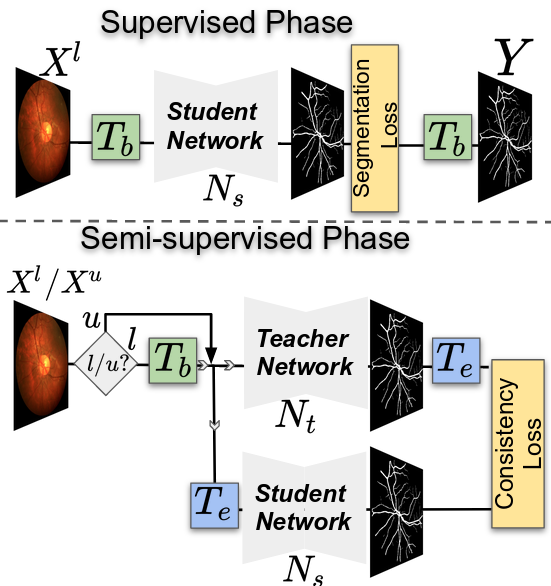}
\includegraphics[width=0.99\linewidth, keepaspectratio]{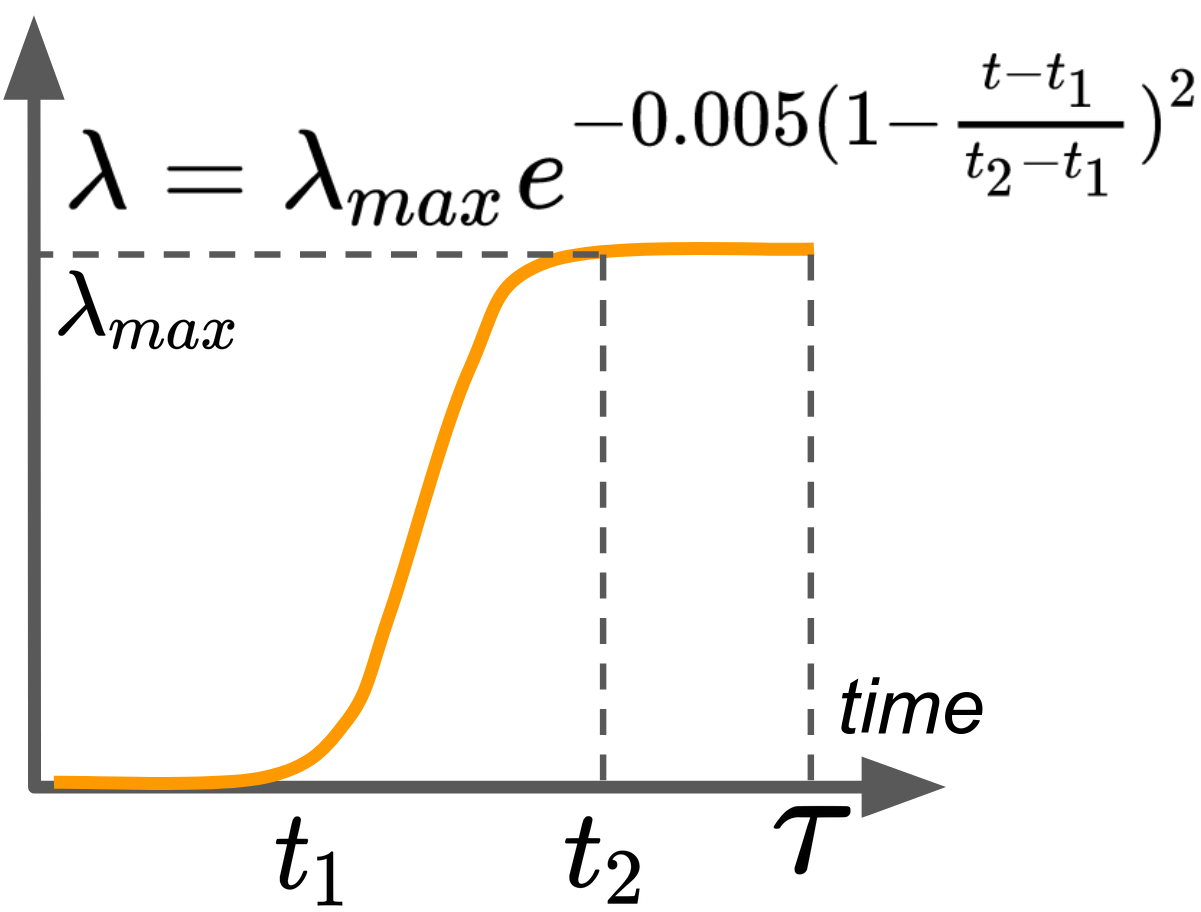}
\end{minipage}
\caption{\small (left) Algorithm 1 shows the pseudocode of our semi-supervised learning method based on extreme consistency. $sDice$ denotes soft dice. (top right) Schematic of the supervised phase (Lines 2--4) and semi-supervised phase (Lines 5--6). (bottom right) Sigmoid ramp-up function used for consistency loss weight (Line 7).}
\label{fig:overview}
\end{figure}

\vspace{4pt}
\noindent{\bf Pseudocode:} \figurename~\ref{fig:overview} presents the pseudocode of our SSL method. In each iteration, a batch of labeled $\{X^l,Y\}$ and unlabeled images $\{X^u\}$ are sampled. During the the supervised phase, the labeled batch is first sent to $T_b$ for mild augmentation, and then to the student network to compute the segmentation loss $L_{seg}=Dice(N_s(T_b(X^l)), T_b(Y))$. For unsupervised learning, we compute the extreme consistency losses $L^l_{ec}$ and $L^u_{ec}$ for both the labeled and unlabeled data. For the former, this is realized by first applying extreme transformations $T_e$ to the batch of images before feeding it to the student network $N_s(T_e(T_b(X^l))$. To generate the reference predictions for the consistency loss, the original labeled batch is sent to the teacher network and the equivalent extreme transformations $T_e(N_t(T_b(X^l)))$ are applied to its predictions. The extreme consistency loss $L^l_{ec}$ is then computed by comparing the two outputs through a soft Dice function (both outputs are probabilistic). The extreme consistency loss for the unlabeled batch $L^u_{ec}$ is computed similarly, with the difference being that no basic transformation $T_b$ is applied to the unlabeled data. This choice is made because the student and teacher networks do not have the ground truth for unlabeled images during training; hence, applying mild transformations may degrade the quality of reference predictions for the unlabeled data. Once the three losses are computed, the student network is updated by minimizing $L_{seg}+\lambda(L^l_{ec}+L^ul_{ec})$, where $\lambda$ is governed by a sigmoid function (\figurename~\ref{fig:overview}). Finally, the teacher network is updated by the moving average filter. We use the teacher network during inference.

\section{Experiments}
\label{sec:exp}

\noindent{\bf {Datasets:}} 
We employ a skin lesion segmentation dataset (ISIC'17: 2,000/150/600) and two fundus image datasets (HRF~\cite{budai2013robust}: 30/6/9 and STARE\cite{hoover2000locating}: 12/4/4) in our experiments (tr/val/test). The relatively small size of fundus datasets is due to the difficulty and high cost associated with vessel annotation, which makes them a practical and meaningful choice for addressing the limited annotation problem. The selected fundus datasets also show great variability in terms of pathologies  as well as variations in scanner devices. For instance, HRF consists of high-quality fundus images acquired by a modern Canon scanner whereas the STARE images are acquired by older scanners and exhibit more pathologies. Therefore, this pair of datasets is suitable for studying the domain shift problem.  The skin dataset, on the other hand, is far larger than the fundus datasets, facilitating the large-scale evaluation of the suggested method. This dataset has also been used as a benchmark by one of the competing methods \cite{li2019transformation}, which allows us to draw more direct comparisons with it. 

\vspace{4pt}
\noindent{\bf Experiment protocols:} For the limited annotation problem, we use all three datasets by dividing their training sets into labeled and unlabeled subsets, resulting in $\mathcal{L}$/$\mathcal{U}$ splits of 50/1,950 for ISIC, 3/27 for HRF, and 2/10 for STARE. For the domain shift problem, we use HRF as the source and STARE as the target dataset; that is, $\mathcal{L}$ and $\mathcal{U}$ consist of the training splits of HRF and STARE, respectively. For realistic evaluation, we choose the best segmentation model based on the validation split of the \emph{source} dataset, and then evaluate the best model on the test split of the \emph{target} dataset. Similar data splits and evaluation strategies have been adopted in previous works.

\vspace{4pt}
\noindent{\bf Competing baselines:} For the limited annotation problem, we compare our method with two SSL methods, \cite{cui2019semi} and \cite{li2019transformation}, which use consistency loss based on noise perturbations and affine transformations, respectively. The primary baselines for the domain shift problem are unsupervised domain adaptation methods suggested in \cite{dong2018unsupervised} and \cite{perone2019unsupervised}. The secondary baselines are the above SSL methods (\cite{cui2019semi} and \cite{li2019transformation}), which can potentially be used for domain adaptation as well.
% We have implemented all three methods based on their corresponding publications. For fair comparisons with the proposed method, Following \cite{oliver2018realistic}, we have placed all implementations in one codebase, so all methods use the same segmentation network and benefit from the same set of data augmentation transformations.

\vspace{4pt}
\noindent{\bf Implementation details:}
For a fair comparison, as suggested in \cite{oliver2018realistic}, we have implemented our method as well as all competing baselines in the same codebase. Thus, all methods benefit from the same data augmentation and further operate on the same segmentation network (exception being \cite{perone2019unsupervised} requiring GroupNorm \cite{wu2018group}). We have also optimized the weight function of the consistency loss for each method and dataset, by evaluating different configurations on the validation split. See our supplementary material for the best hyper-parameters.

%That said, we observed that the hyper-parameters found on the HRF dataset generalized fairly well to all other datasets, with the exception of \cite{cui2019semi}, which required extensive hyper-parameter tuning. 
% We have also used the Adam optimizer for all methods used for comparison, controlling for the choice of the optimizer.

\vspace{4pt}
\noindent{\bf Segmentation Architecture:}
We use a U-Net with 4 blocks in the encoder and decoder. Each block consists of 2 Conv-Bn-Relu followed by a 2x2 max-pooling layer in the encoder and an up-sampling layer based on bi-linear interpolation in the decoder. To avoid overfitting, we place a dropout layer with 50\% drop rate right after each encoder block. We use the Dice loss during training. 
%To generate segmentation results, we apply a 1x1 convolution followed by a softmax function.  We use a Dice-based segmentation loss. 
Being optimized on the validation set, this U-Net performs at par with the sophisticated, heavy-weight UNet++ \cite{zhou2018unet++} (Table~\ref{tab:ssl_results}). This observation is in line with similar findings in \cite{isensee18}. We thus use the U-Net for the remaining experiments. 

\begin{table}[t]
\centering
\caption{\small 10-trial comparison between our method and the competing baselines for (a) limited annotation and (b) the domain shift problem. In both tables, the goal is to bridge the performance gap between the lower bound (LB) and upper bound (UB) performance levels. To mimic limited annotation, we have divided the training split $D_{tr}$ of each of the datasets into a small labeled subset $\mathcal{L}$ and a large subset $\mathcal{U}$ whose annotations are removed. As described in the text, the sizes of $D_{tr}$, $\mathcal{L}$, $\mathcal{U}$ differ from one dataset to another. For domain adaptation, we use HRF as the source and STARE as the target dataset. We use the training split of STARE after removing the labels (unsupervised), denoted as $\cancel{D}_{tr}^S$. $p$-values compare our method with the second best.}
\label{tab:ssl_results}
\vspace{4pt}
\resizebox{0.51\textwidth}{!}{%
\begin{tabular}{lcccc}
\multicolumn{5}{l}{ {\Large {\bf (a)} Limited annotation} }\\ 
\toprule
\multicolumn{1}{c}{Method} &  & \multicolumn{3}{c}{Dice: $\mu$($\sigma$)}  \\ 
\midrule
\multicolumn{1}{c}{} & Tr set & \multicolumn{1}{c}{STARE} & \multicolumn{1}{c}{HRF} & \multicolumn{1}{c}{ISIC'17} \\ 
\hline
LB (U-Net) & $\mathcal{L}$ & 0.806(0.003) & 0.807(0.012) & 0.775(0.005)   \\ 
LB (UNet++) & $\mathcal{L}$ & 0.797(0.007) & 0.812(0.003) & 0.781(0.004)   \\ 
Li et al. \cite{li2019transformation} & $\mathcal{L}$+$\mathcal{U}$ & 0.809(0.006) & 0.811(0.006) & {\bf 0.797(0.005)}  \\ 
Cui et al. \cite{cui2019semi} & $\mathcal{L}$+$\mathcal{U}$ & {\bf 0.814(0.013)} & {\bf 0.813(0.005)} & 0.774(0.016)    \\ 
\rowcolor{orange!50}
Proposed & $\mathcal{L}$+$\mathcal{U}$ & {\bf 0.824(0.007)} & {\bf 0.828(0.002)} & {\bf 0.806(0.005)}   \\ 
\midrule
UB (U-Net) & $D_{tr}$ & 0.836(0.006) & 0.838(0.001) & 0.840(0.001)   \\ 
UB (UNet++) & $D_{tr}$ & 0.827(0.005) & 0.839(0.003) & 0.832(0.002)   \\ 
\midrule
\multicolumn{2}{l}{{\large $p$-value}} & 0.079 & $p<$0.001 & $p<$0.05   \\ 
\bottomrule
\end{tabular}} 
% \quad
\resizebox{0.48\textwidth}{!}{%
~~~\begin{tabular}{lcccc}
\multicolumn{5}{l}{ {\large {\bf (b)} Domain shift: {\bf H}RF $\rightarrow${\bf S}TARE} }\\ 
\toprule
\multicolumn{1}{c}{Method} & Tr set & \multicolumn{1}{c}{Val set} & \multicolumn{1}{c}{Test Set} & \multicolumn{1}{c}{Dice: $\mu$($\sigma$)} \\ 
\midrule
LB (U-Net) & $D_{tr}^{H}$ & $D_{val}^{H}$ & $D_{test}^{S}$ & 0.678(0.010)    \\ 
LB (UNet++) & $D_{tr}^{H}$ & $D_{val}^{H}$ & $D_{test}^{S}$ & 0.655(0.014)    \\ 
Li et al. \cite{li2019transformation} & $D_{tr}^{H}$+$\cancel{D}_{tr}^S$ & $D_{val}^{H}$ & $D_{test}^{S}$ & 0.712(0.005)    \\ 
Cui et al. \cite{cui2019semi} & $D_{tr}^{H}$+$\cancel{D}_{tr}^S$ & $D_{val}^{H}$ & $D_{test}^{S}$ & 0.704(0.012)    \\ 
Dong et al. \cite{dong2018unsupervised} & $D_{tr}^{H}$+$\cancel{D}_{tr}^S$ & $D_{val}^{H}$ & $D_{test}^{S}$ & 0.668(0.017)    \\ 
Perone et al. \cite{perone2019unsupervised} & $D_{tr}^{H}$+$\cancel{D}_{tr}^S$ & $D_{val}^{H}$ & $D_{test}^{S}$ & {\bf 0.723(0.012)}    \\ 
\rowcolor{orange!50}
Proposed & $D_{tr}^{H}$+$\cancel{D}_{tr}^S$ & $D_{val}^{H}$ & $D_{test}^{S}$ & {\bf 0.736(0.004)}    \\ 
\midrule
UB (U-Net) & $D_{tr}^{S}$ & $D_{val}^{S}$ & $D_{test}^{S}$ & 0.833(0.005)    \\ 
UB (UNet++) & $D_{tr}^{S}$ & $D_{val}^{S}$ & $D_{test}^{S}$ & 0.827(0.005)    \\ 
\midrule
\multicolumn{1}{l}{{\large $p$-value}} & &  &  & $p<$0.05  \\ 
\bottomrule
\end{tabular}} 
\end{table}

\vspace{4pt}
\noindent{\bf Results for limited annotation:} Table~\ref{tab:ssl_results}(a) summarizes the results. Each model is trained and evaluated in 10 trials. According to our statistical analyses, our method has significantly outperformed the best-performing SSL method for HRF and ISIC'17. Notably, our results for \cite{li2019transformation} show a $\sim$2-point increase over the lower bound U-Net in the ISIC'17 dataset, which is consistent with the original publication where the same dataset is used for evaluation. Our method has further outperformed the second-best method in STARE in both the mean and standard deviation.

\vspace{4pt}
\noindent{\bf Results for domain shift:} The results are summarized in Table~\ref{tab:ssl_results}(b). Each model is trained and evaluated in 10 trials. Our method achieves a $\sim$6-point increase in Dice over the lower bound performance (U-Net), on average $\sim$3-point increase over the SSL methods, and a 1-point increase over the best performing domain adaptation method. All improvements are statistically significant. These results demonstrate the ability of our model to handle domain shifts.

\vspace{4pt}
\noindent{\bf Ablation study:} Table~\ref{tab:ablation} summarizes the results of our ablation study conducted on the HRF dataset. First, we studied the necessity of diverse transformations by using one transformation type at a time, observing significant drops in Dice ($p<0.05$) compared to using diverse transformations. We further compared diverse transformation with and without our image mixing method (the last two rows in Table~\ref{tab:ablation}(a)), observing a substantial decrease in average Dice without image mixing. It is also clear that image mixing significantly reduces the standard deviation of Dice, which translates to far more stable training sessions and reproducible models. Second, we removed the exclusion principle, applying the diverse transformations to the inputs of both student and teacher networks. This change led to a significant drop in Dice (3\%), degrading the performance to that of the lower bound supervised model, motivating the the need for the exclusive transformations. Third, we excluded unlabeled data, computing segmentation and consistency losses using only labeled data. This configuration led to 1.4\% drop in Dice.  Fourth, we studied the effect of replacing the teacher  with the student network, which is known as the $\Pi$ model \cite{laine2016temporal}, observing a 0.6\% drop in Dice. Fifth, we trained the lower bound U-Net with extreme data augmentation ($T_e$), observing no significant gains, which rules out the possibility that the gains by our SSL method are due to extensive data augmentation. These experiments underline the importance of each individual component of our method.

\begin{table}[t]
\centering
\caption{\small 10-trial ablation study for HRF. We have split the results into three tables: (a) the necessity of diverse transformations and our image mixing method (last two rows); (b) impact of exclusive transformations (+Exclusive), use of unlabeled data (+$\mathcal{U}$), and the necessity of a mean teacher network (+Teacher); (c) whether a supervised model with extreme augmentation is as effective as our method. 
%All components above contribute to the final performance by either increasing the mean or decreasing  the standard deviation of the Dice metric.
}
\label{tab:ablation}
\resizebox{0.48\textwidth}{!}{%
\vspace{4pt}
\begin{tabular}{ccccc}
\multicolumn{5}{l}{  {\bf (a)} Diverse transformations} \\ 
\toprule
+Inten. & +Geo. & +Mixing & Dice: $\mu$($\sigma$) & $\Delta$Dice(\%)\\ 
\midrule
\checkmark& \ding{56} & \ding{56} & 0.817(0.002)& -1.3\%\\ 
\ding{56}& \checkmark & \ding{56} & 0.814(0.020)& -1.6\%\\ 
\ding{56}& \ding{56} & \checkmark & 0.817(0.008)& -1.3\%\\ 
\checkmark& \checkmark & \ding{56} & 0.818(0.019)& -1.3\%\\ 
\rowcolor{orange!50}
\checkmark& \checkmark & \checkmark & {\bf 0.828(0.002)}& 0.0\%\\ 
\bottomrule
\end{tabular}} 
\resizebox{0.48\textwidth}{!}{%
\vspace{2pt}
~~~\begin{tabular}{ccccc}
\multicolumn{5}{l}{ \large  {\bf (b)} Exclusive transformations + others} \\ 
\toprule
+Exclusive & +$\mathcal{U}$ & +Teacher & Dice: $\mu$($\sigma$) & $\Delta$Dice(\%)\\ 
\midrule
\ding{56} &\checkmark & \checkmark & 0.803(0.005)& -3.0\%\\ 
\checkmark & \ding{56}  & \checkmark & 0.816(0.002)& -1.4\%\\ 
\checkmark & \checkmark & \ding{56} & 0.823(0.004)& -0.6\%\\ 
\rowcolor{orange!50}
\checkmark& \checkmark & \checkmark & {\bf 0.828(0.002)} & 0.0\%\\ 
\bottomrule
\noalign{\vskip 10pt} 
\multicolumn{5}{l}{ \large {\bf (c)} Supervised model with extreme augmentation} \\ 
\toprule
\multicolumn{2}{l}{model}  & \large $T_b$ & \large $T_e$ & \large Dice\\ 
\midrule
\multicolumn{2}{l}{Sup. U-Net} & \checkmark & & 0.807(0.012)  \\ 
\multicolumn{2}{l}{Sup. U-Net} & \checkmark & \checkmark & 0.801(0.007)  \\ 
\rowcolor{orange!50}
\multicolumn{2}{l}{Ours: SSL U-Net} & \checkmark & \checkmark & 0.828(0.002)  \\ 
\bottomrule
\end{tabular}} 
\vspace{-10pt}
\end{table} 

\section{Conclusion}

We presented Extreme Consistency as a simple yet effective semi-supervised learning framework. Although our method is intuitive and simple to implement, it demonstrates a substantial performance gain over several baselines across multiple datasets in handling both limited annotation and domain shift problems. Through extensive ablation studies, we further demonstrated that each component of our method contributes to its success. While our empirical results  are strong, surpassing state-of-the-art performances in most of the applications, important future work will include the exploration of applications of extreme consistency beyond image segmentation and the currently used datasets. 

% \section{TODO}
% \begin{itemize}
%     \item Emphasize on how some methods perform well on 1 dataset but then yielding mixed results across multiple datasets. For instance, ours outperforms on HRF but the gap is bridged on stare. This explains why PBR under-performs on stare.
    
%     \item unlike PBR and lequan, we do not degrade the reference image passed to the teacher network
% \end{itemize}

\bibliographystyle{splncs04}
\bibliography{refs}

\newpage
\input{SM.tex}

\end{document}

%% file: SM.tex
\section*{Supplementary Material}
\setcounter{figure}{0}
\setcounter{table}{0}

\begin{figure}[h]
\centering
\includegraphics[width=0.85\linewidth, keepaspectratio]{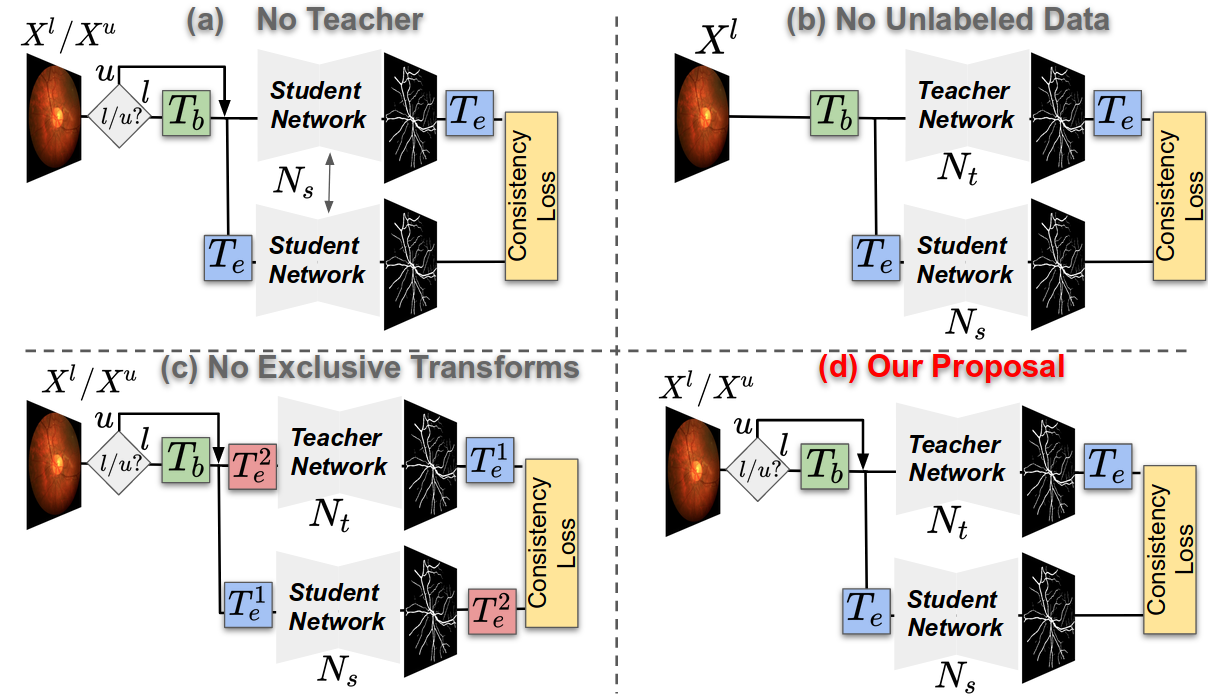}
    \caption{\small 
    Different variants of our method used in the ablation study. The supervised phase is the same for all variants; hence, not shown here. Transformation blocks with the same color perform the same transformation on the input data. (a) the teacher network is replaced with the student network. (b) consistency loss is computed for only labeled data. (c) Diverse transformation are also applied to the teacher. (d) The proposed architecture, which uses teacher, unlabeled data, and exclusive perturbations.}
    \label{fig:ablation}
\end{figure}

\begin{figure}[h!]
\centering
% \includegraphics[width=0.24\linewidth, keepaspectratio]{pics/SM/qual_res/ISIC_0015386.jpg}
% \includegraphics[width=0.24\linewidth, keepaspectratio]{pics/SM/qual_res/ISIC_0015386.jpg}
% \includegraphics[width=0.24\linewidth, keepaspectratio]{pics/SM/qual_res/ISIC_0015386_low.jpg}
% \includegraphics[width=0.24\linewidth, keepaspectratio]{pics/SM/qual_res/ISIC_0015386_high.jpg}
% \includegraphics[width=0.24\linewidth, keepaspectratio]{pics/SM/qual_res/ISIC_0015386_segmentation.png}
% \\
\begin{minipage}[t]{.5\textwidth}
  \vspace{0pt}  
\includegraphics[width=0.99\linewidth, keepaspectratio]{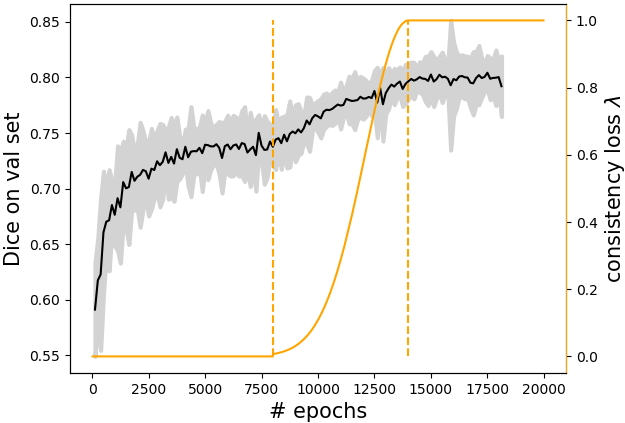}
\end{minipage}
\hfill
\begin{minipage}[t]{.45\textwidth}
  \vspace{0pt}  
\includegraphics[width=0.4\linewidth, keepaspectratio]{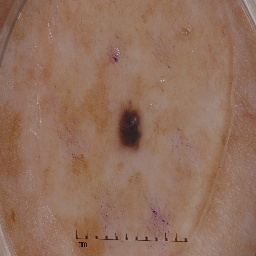}
\includegraphics[width=0.4\linewidth, keepaspectratio]{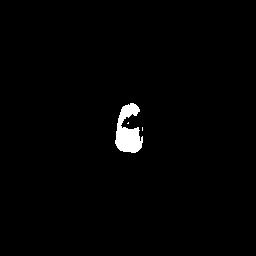}\\
\includegraphics[width=0.4\linewidth, keepaspectratio]{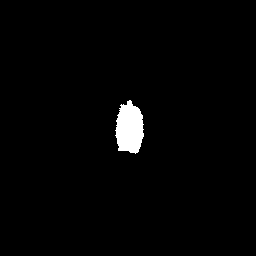}
\includegraphics[width=0.4\linewidth, keepaspectratio]{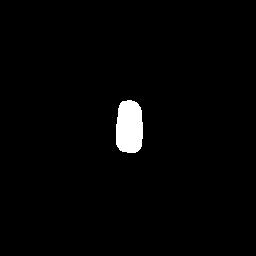}
\end{minipage}
    \caption{\small (left) Effect of extreme consistency (EC) loss on the convergence of the teacher network. ISIC'17 is used for training and evaluation. The shaded plot indicates the 95\% confidence interval of 10 trials. The initiation of EC at 8,000 epochs starts a new increasing trend in the validation Dice, which results in a marked improvement over the baseline supervised model. (right) clockwise: original image, segmentation without EC, segmentation with EC, and ground truth (GT). 
    % Due to the page limit on supplementary material, no more examples can be included.
   }
    \label{fig:ec_imporovement}
\end{figure}

% \usepackage{multirow}
% \usepackage{graphicx}
\begin{table}[h!]
\caption{Hyper-parameters for our method and other consistency-based methods used for comparison. For each method and dataset, hyper-parameters are tuned on the corresponding validation set. We use a moving average decay of $\alpha$= 0.999 and further employ the Adam optimizer with a learning rate of 0.001. }
\label{tab:my-table}
\centering
\resizebox{0.8\textwidth}{!}{%
\begin{tabular}{c|ccccccccc|cccc}
\hline
\multirow{3}{*}{Hyper-parameters} & \multicolumn{9}{c|}{\textbf{Limited annotations}} & \multicolumn{4}{c}{\textbf{Domain shift}} \\ \cline{2-14} 
 & \multicolumn{3}{c}{ISIC'17} & \multicolumn{3}{c}{HRF} & \multicolumn{3}{c|}{STARE} & \multicolumn{4}{c}{HRF $\rightarrow$ STARE} \\ \cline{2-14} 
 & \cite{li2019transformation} & \cite{cui2019semi} & Ours & \cite{li2019transformation} & \cite{cui2019semi} & Ours & \cite{li2019transformation} & \cite{cui2019semi} & Ours & \cite{li2019transformation} & \cite{cui2019semi} & \cite{perone2019unsupervised} & Ours \\ \hline
t1 & 8k & 8k & 8k & 8k & 8k & 8k & 2k & 8k & 8k & 4k & 4k & 2k & 0 \\ \hline
t2 & 14k & 14k & 14k & 14k & 14k & 14k & 6k & 14k & 14k & 8k & 8k & 6k & 6k \\ \hline
$\lambda_{max}$ & 1.0 & 0.1 & 1.0 & 1.0 & 1.0 & 1.0 & 0.1 & 1.0 & 1.0 & 0.1 & 0.1 & 1.0 & 0.1 \\ \hline
\end{tabular}%
}
\end{table}
\vspace{-1.0cm}
\begin{table}[h!]
\caption{List of basic and extreme augmentations along with their corresponding parameters. Extreme augmentation consists of a random selection of 1 geometric and 3 intensity transformations followed by a 50\% chance of applying image mixing. We use kornia~\cite{eriba2019kornia,Arraiy2018} to implement some of the transformations below.}
\label{tab:adv-augments}
\vspace{2pt}
\resizebox{\textwidth}{!}{%
\begin{tabular}{lllll}
\hline
\textbf{Augmentation} &  & \textbf{Description} & \textbf{Parameter} & \textbf{Values} \\ \hline
\multicolumn{5}{c}{\textbf{BASIC AUGMENTATIONS}} \\ \hline
Saturation Jitter &  & Scale the saturation channel by a random constant & Saturation factor & U(0.9, 1.1) \\
Hue Jitter &  & Shifts the hue channel by a random constant & Hue factor & U(-0.1, 0.1) \\
Flip &  & Flips the image along the vertical or horizontal axis &  &  \\
Channel Shuffle &  & Randomizes the order of image channels &  &  \\
Affine &  & Applies affine transformation consisting of rotation, scaling and translation & \begin{tabular}[c]{@{}l@{}}Rotation\\ Scale\\ Translation\end{tabular} & \begin{tabular}[c]{@{}l@{}}U(-10,10)\\ U(0.75,1.25)\\ U(-32,32)\end{tabular} \\ \hline
\multicolumn{5}{c}{\textbf{EXTREME AUGMENTATIONS}} \\ \hline
\textbf{Intensity} &  &  &  &  \\ \hline
Autocontrast &  & \begin{tabular}[c]{@{}l@{}}Maximizes the image contrast by stretching the min and max\\  of each image channel and then blends it with the original image.\end{tabular} & Blending ratio & U(0, 1) \\
Brightness &  & \begin{tabular}[c]{@{}l@{}}Adjusts the image brightness. An enhancement factor of 0 \\ gives a black image and 1 gives the original image.\end{tabular} & Enhancement factor & U(0.5, 1) \\
Color &  & \begin{tabular}[c]{@{}l@{}}Adjusts the color balance of an image. An enhancement factor of\\ 0 gives a black-and-white image while 1 gives the original image.\end{tabular} & Enhancement factor & U(0, 1) \\
Contrast &  & \begin{tabular}[c]{@{}l@{}}Decreases the image contrast. An enhancement factor of\\ 0 gives a solid grey image and 1 gives the original image.\end{tabular} & Enhancement factor & U(0.5, 1) \\
Equalize &  & Equalizes image channels and then blends it with the original image. & Blending ratio & U(0, 1) \\
Invert &  & Inverts all pixel intensities and then blends it with the original image. & Blending ratio & \begin{tabular}[c]{@{}l@{}}U(0,0.25) or\\  U(0.75,1)\end{tabular} \\
Solarize &  & Inverts all pixels whose intensities are above a threshold & Threshold & U(0.3, 1) \\
Posterize &  & Reduces number of bits for each color channel & Number of bits & U(4,8) \\ \hline
\textbf{Geometric} &  &  &  &  \\ \hline
Rotation &  & Rotates the image & Extent & U(-10, 10) \\
Translation &  & Translates the image horizontally or vertically & Extent & U(-32, 32) \\
Scale &  & Scales the image & Scale factor & U(0.75, 1.25) \\
Flip &  & Randomly flips along the vertical or horizontal axis &  &  \\ \hline
\textbf{Image Mixing} &  &  &  &  \\ \hline
Cross-dataset image manipulation &  & \begin{tabular}[c]{@{}l@{}}Mixes two images by copying a square patch from one to the other.\\ The patch location is random and size is a random fraction of the image.\end{tabular} & Size ratio & U(0, 0.5) \\ \hline
\end{tabular}%
}
\end{table}

\begin{figure}[H]
\centering
  \includegraphics[width=0.12\linewidth, keepaspectratio]{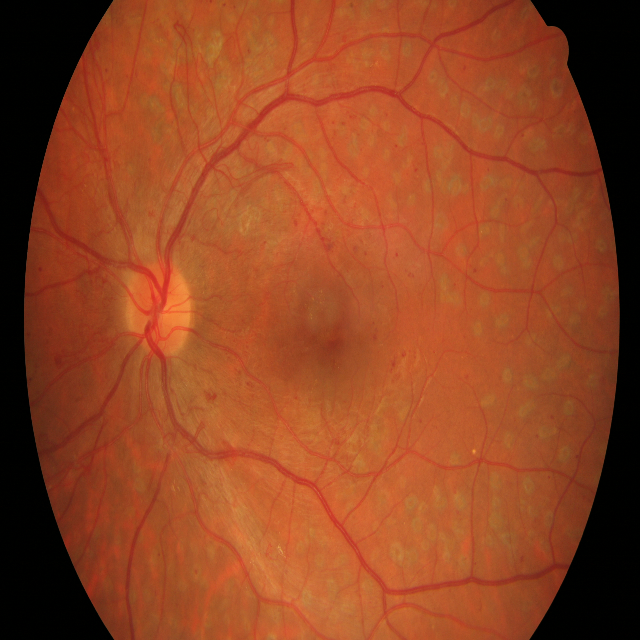}
\hspace{-7pt}
\includegraphics[width=0.12\linewidth, keepaspectratio]{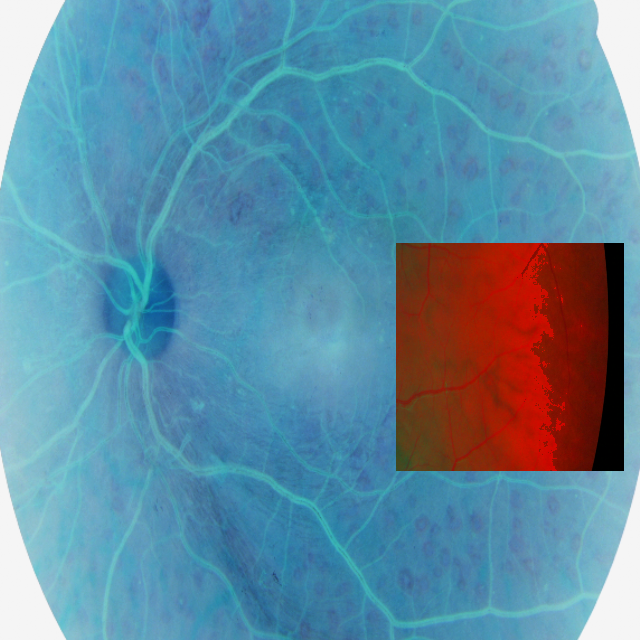}
\includegraphics[width=0.12\linewidth, keepaspectratio]{pics/SM/extreme_ex/orig3.png}
\hspace{-7pt}
\includegraphics[width=0.12\linewidth, keepaspectratio]{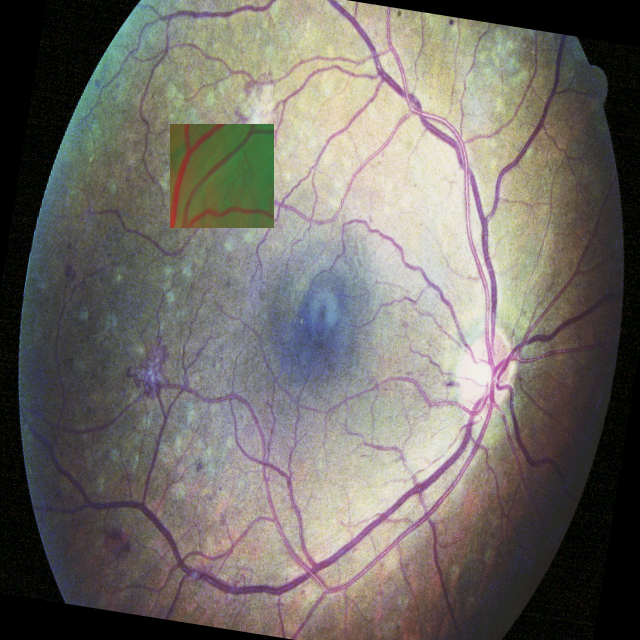}
\includegraphics[width=0.12\linewidth, keepaspectratio]{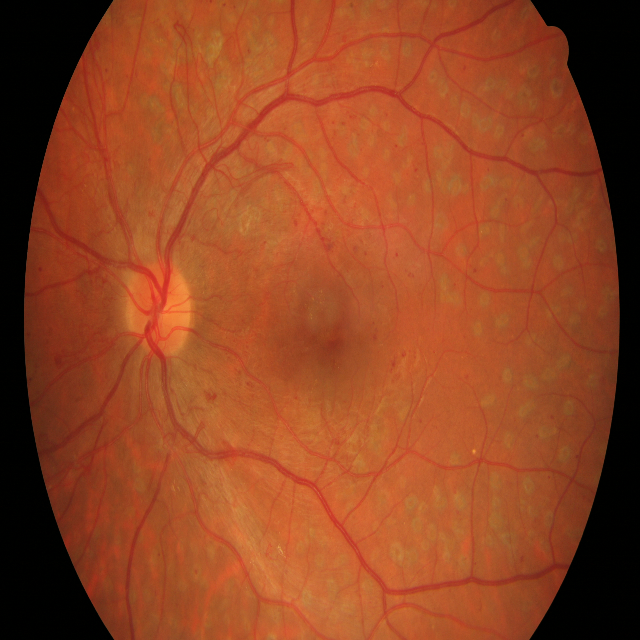}
\hspace{-7pt}
\includegraphics[width=0.12\linewidth, keepaspectratio]{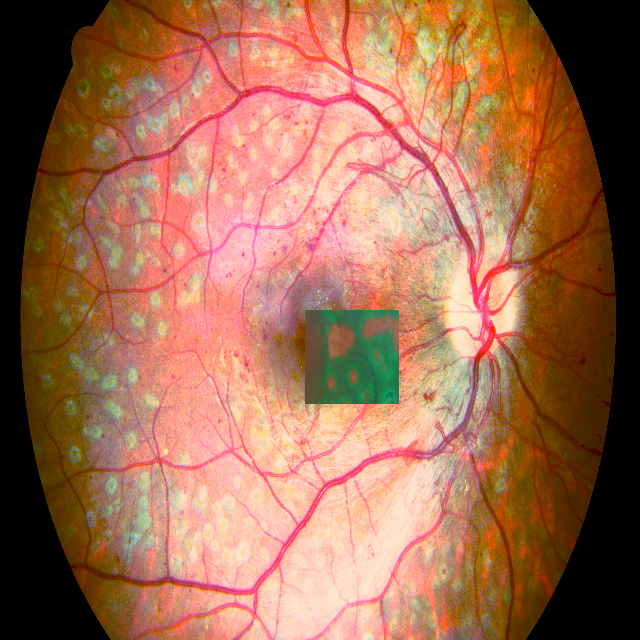}
\includegraphics[width=0.12\linewidth, keepaspectratio]{pics/SM/extreme_ex/orig8.png}
\hspace{-7pt}
\includegraphics[width=0.12\linewidth, keepaspectratio]{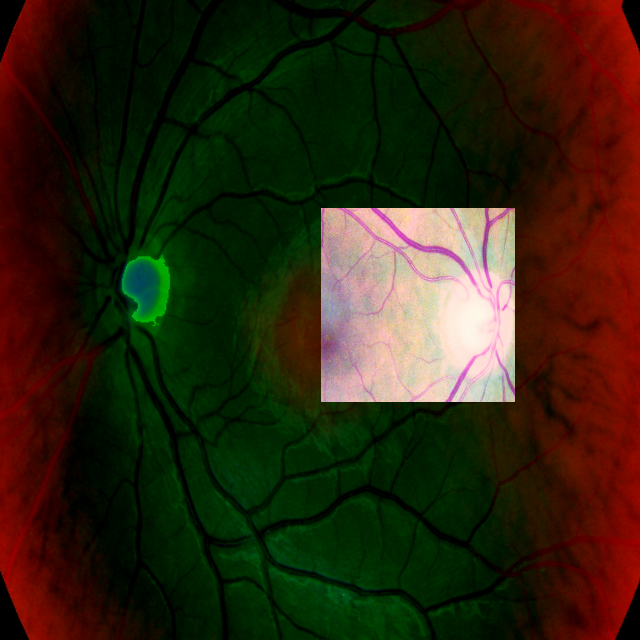}
    \caption{\small Examples of original images and their extremely augmented versions, where the latter is produced using intensity, geometric, and mixing transformations.}
    \label{fig:extreme_transforms}
\end{figure}